\documentclass[sigconf,natbib]{acmart}

\usepackage[ruled,vlined]{algorithm2e}

\usepackage{graphicx}
\usepackage{url}
\usepackage{multirow}
\usepackage{comment}
\usepackage{pifont}
\usepackage{mathtools}
\usepackage{xcolor}
\usepackage{enumitem}
\usepackage{marvosym}
\usepackage{textcomp}
\usepackage{array}
\usepackage{booktabs}
\usepackage{multicol}
\usepackage{balance}
\usepackage{amsmath}
\usepackage{xcolor}
\usepackage{calc}
\usepackage{diagbox}
\usepackage{smile} 
\usepackage{xcolor,colortbl}
\usepackage{subfig}
\usepackage{amsfonts}
\usepackage{bbm}

\copyrightyear{2023}
\acmYear{2023}
\setcopyright{acmlicensed}\acmConference[SIGIR '23]{Proceedings of the 46th International ACM SIGIR Conference on Research and Development in Information Retrieval}{July 23--27, 2023}{Taipei, Taiwan}
\acmBooktitle{Proceedings of the 46th International ACM SIGIR Conference on Research and Development in Information Retrieval (SIGIR '23), July 23--27, 2023, Taipei, Taiwan}
\acmPrice{15.00}
\acmDOI{10.1145/3539618.3592085}
\acmISBN{978-1-4503-9408-6/23/07}

\definecolor{darkpink}{rgb}{0.91, 0.33, 0.5}

\newcommand{\blue}[1]{{\color{blue}#1}}

\makeatletter
\newcommand{\thickhline}{%
    \noalign {\ifnum 0=`}\fi \hrule height 1pt
    \futurelet \reserved@a \@xhline
}

\usepackage{xspace}
\newcommand{\ours}{\textsc{WanDeR}\xspace}

\acmSubmissionID{4704}

\begin{document}

\title{Weakly-Supervised Scientific Document Classification via Retrieval-Augmented Multi-Stage Training}

\author{Ran Xu}
\authornote{Ran and Yue contributed equally to this research.}
\affiliation{%
  \institution{Emory University}
  \city{Atlanta}
  \state{GA}
  \country{USA}
}
\email{ran.xu@emory.edu}
\author{Yue Yu}
\authornotemark[1]
\affiliation{%
  \institution{Georgia Institute of Technology}
  \city{Atlanta}
  \state{GA}
  \country{USA}
}
\email{yueyu@gatech.edu}
\author{Joyce C. Ho}
\affiliation{%
  \institution{Emory University}
  \city{Atlanta}
  \state{GA}
  \country{USA}
}
\email{joyce.c.ho@emory.edu}
\author{Carl Yang}
\authornote{Corresponding Author.}
\affiliation{%
  \institution{Emory University}
  \city{Atlanta}
  \state{GA}
  \country{USA}
}
\email{j.carlyang@emory.edu}

\begin{abstract}
Scientific document classification is a critical task for a wide range of applications, but the cost of obtaining massive amounts of human-labeled data can be prohibitive.  
To address this challenge, we propose a weakly-supervised approach for scientific document classification using label names only. 
In scientific domains, label names often include domain-specific concepts that may not appear in the document corpus, making it difficult to match labels and documents precisely. 
To tackle this issue, we propose {\ours}, which leverages \emph{dense retrieval} to perform matching in the embedding space to capture the semantics of label names. 
We further design the label name expansion module to enrich the label name representations. 
Lastly, a self-training step is used to refine the predictions. The experiments on three datasets show that {\ours} outperforms the best baseline by 11.9\% on average. Our code will be published at \url{https://github.com/ritaranx/wander}.

\vspace{-0.5ex}
\end{abstract}
\maketitle
\vspace{-1ex}
\section{Introduction}
\vspace{-0.3ex}
Scientific document classification aims to assign scientific literature to pre-defined categories, supporting various applications~\cite{cohan2020specter,xie2021learning,zhuang-etal-2022-resel}. Recently, pretrained language models (PTLMs) have demonstrated impressive performance in document classification~\cite{devlin2019bert, beltagy2019scibert}. However, they often  require a large number of annotations for fine-tuning, which restricts their deployment in real-world applications.
While practitioners cannot afford to label many
documents, it is often easier for them to provide category-descriptive label names as \emph{weak supervision} for each class~\cite{meng2018weakly,zhang2wrench,prboost,zhang2022adaptive}.
Motivated by this,  we focus on scientific document classification under the setting where only the label name for each class as well as the unlabeled corpus are available~\cite{lotclass}. This task is challenging as the label names can be short and succinct, often containing a few words only. How to mine class-relevant knowledge with weak supervision is nontrivial.
 \begin{figure}
	\centering
	\includegraphics[width=0.99\linewidth]{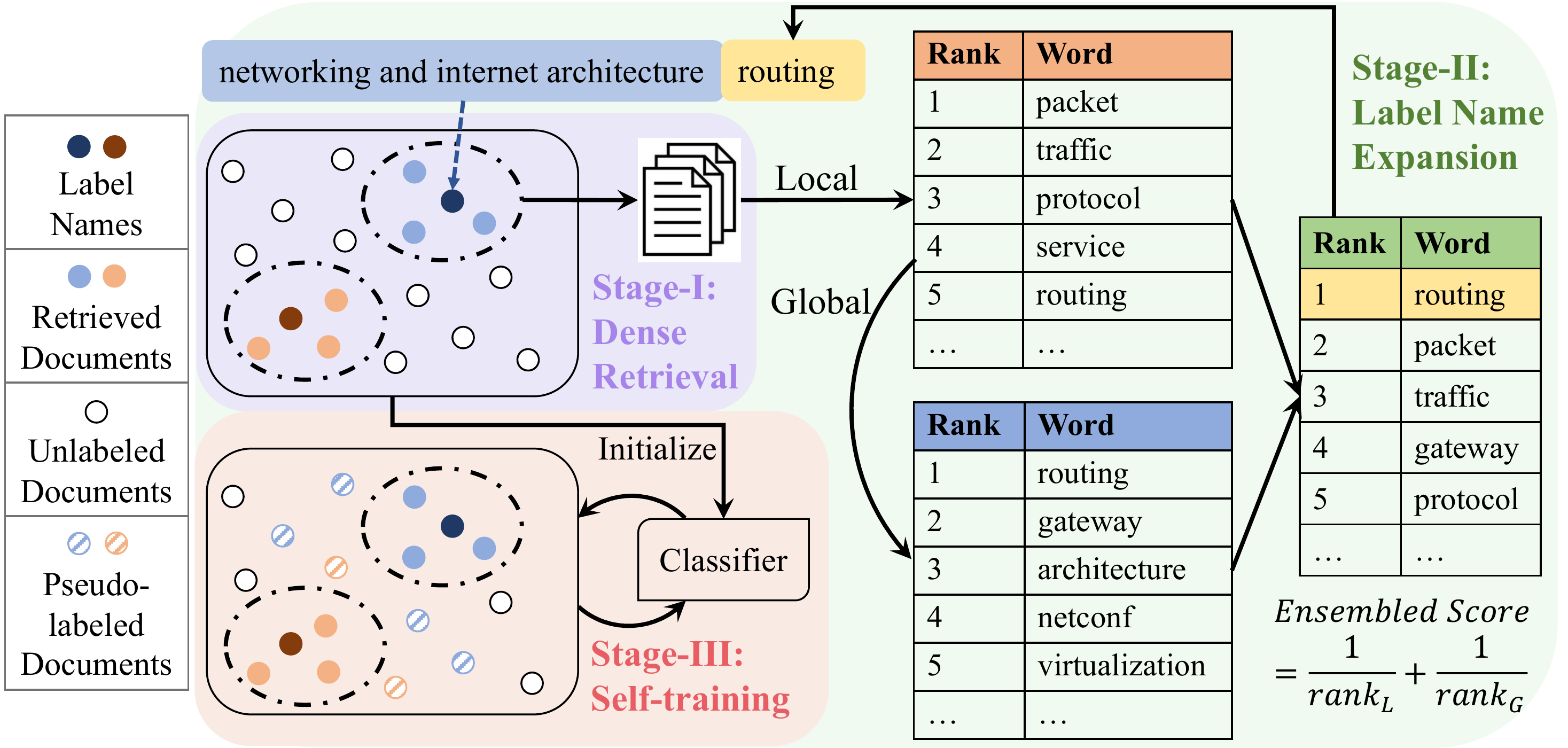}
	\caption{Framework of {\ours}.}
\label{fig:framework}
\end{figure}

There exist plenty of studies on automatic document categorization using class-relevant keywords only \cite{meng2018weakly,lotclass,xclass,yu-etal-2021-fine,wang2023benchmark}. 
These methods first leverage the keywords as input to extract relevant documents from the unlabeled data with hard matching. 
Then, they  enlarge the set of keywords with masked language modeling~\cite{lotclass} or embedding similarity~\cite{xclass}, and use it to generate pseudo labels for unlabeled data.  
Although these methods achieve competitive performance, they mainly focus on tasks from \emph{general domains} such as news and reviews.  
For these tasks, the keywords can be commonly used words (e.g. `\texttt{Good}/\texttt{Bad}' for movie reviews), and they can be matched with many examples from the unlabeled corpus.  
However, for scientific documents, the label names can either be too domain-specific, or contain multiple concepts~\cite{zhang2022seed, cui2022can}.
As a result, they often have limited coverage over the  corpus, which causes performance degradation when directly applying prior techniques on weakly-supervised learning to the scientific domain (Sec.~\ref{sec:challenge}). 

In this work, we propose {\ours} ({\textbf{W}e\textbf{a}kly-supervised Scie\textbf{n}tific Text Classification using \textbf{De}nse \textbf{R}etrieval}), a multi-stage training framework for weakly supervised text classification using dense retrieval (DR), as shown in Figure~\ref{fig:framework}. In DR, both queries and documents are represented as dense vectors, and the relevance between them is calculated via similarity metrics (e.g. dot product)~\cite{dpr}.
This makes DR an ideal choice to tackle the above challenges, as it captures the semantics for different classes and circumvents the mismatch issue since some label names never appear in the corpus. 
To incorporate DR into the framework, we regard label names as queries, and retrieve the most relevant documents from the unlabeled corpus for each class (\textbf{Stage-I}, Sec. \ref{sec:stage1}).
In this way, we create an initial set of pseudo-labeled documents, which can be used to fine-tune the PTLM for the target  task. 

Although Stage-I is able to extract relevant documents, their performance can be less satisfactory as label names are insufficient to capture all the class-specific information. To overcome this drawback, in \textbf{Stage-II}, we expand the label names with the extracted keywords using local and global information (Sec. \ref{sec:stage2}). Specifically, we first adopt the TF-IDF algorithm~\cite{grootendorst2022bertopic} on the 
retrieved documents to select the top-ranked words from the local corpus. 
Besides, as the PTLM captures the generic linguistic knowledge, we use it to calculate the embedding similarity between the candidate words and the label names as the global score. 
The local and global information is connected via an ensemble ranking module, and we augment the label name for each class by selecting the word with the highest score. 
The above expansion step is repeated multiple times to enrich the query~\cite{diaz2016query} and help the DR model retrieve more relevant documents from the corpus. 

To leverage all unlabeled data to further improve the performance, an additional step is to harvest
\emph{self-training}~\cite{lotclass,xu2023neighborhood} (\textbf{Stage-III}, Sec.~\ref{sec:stage3}) to refine
the PTLM classifier by bootstrapping over high-confident examples and improve its generalization ability.  

We verify the effectiveness of {\ours} by conducting experiments on three datasets and show that our model outperforms the previous weakly-supervised approaches by a large margin. 
Our analysis further confirms the advantage of leveraging dense retrieval for tackling the limited coverage issue of label names as well as the efficacy of multi-stage training for improving performance.
  
\vspace{-1ex}
\section{Preliminaries}
\subsection{Problem Definition}
Our weakly-supervised scientific document classification with $C$ classes is defined as follows. The input is a training corpus $\mathcal{X}=\left\{d_1, d_2, \ldots, d_{|\mathcal{X}|}\right\}$ of documents without any labels. In addition, for each class $c$ $(1\leq c \leq C)$, a label-specific name $w_{c}$ is given, which consists of one or a few words.
We aim to learn a classifier $f(x;\theta): \mathcal{X} \rightarrow \mathcal{Y}$. Here $\mathcal{X}$ denotes all samples and $\mathcal{Y}=\{1,2,\cdots,C\}$ is the label set.
While there exist works on multi-label classification or metadata-aware classification~\cite{ganguly2017paper2vec,zhang2023effect}, we focus on the basic setting by assuming (1) each document only belongs to one category and (2) no other metadata information is available. 
\vspace{-1ex}
\subsection{Challenges for Scientific Text Classification}
\label{sec:challenge}
\begin{figure}
	\centering
	\vspace{-2ex}
	\subfloat[Prec./Coverage (in \%).]{
		\includegraphics[width=0.31\linewidth]{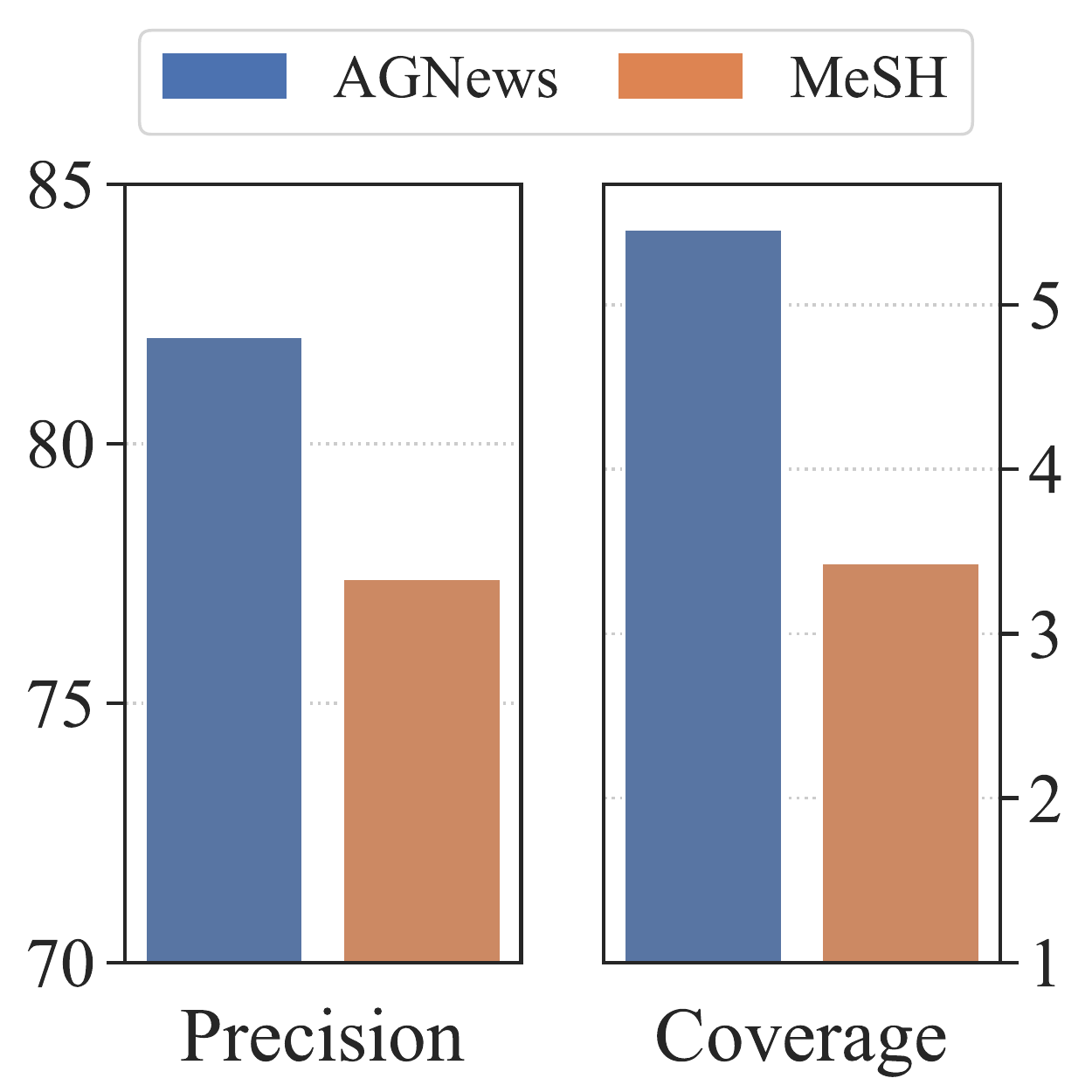}
		\label{fig:precision_recall}
	} 
         \hspace{-1.5ex}
	\subfloat[Per-class Coverage.]{
		\includegraphics[width=0.31\linewidth]{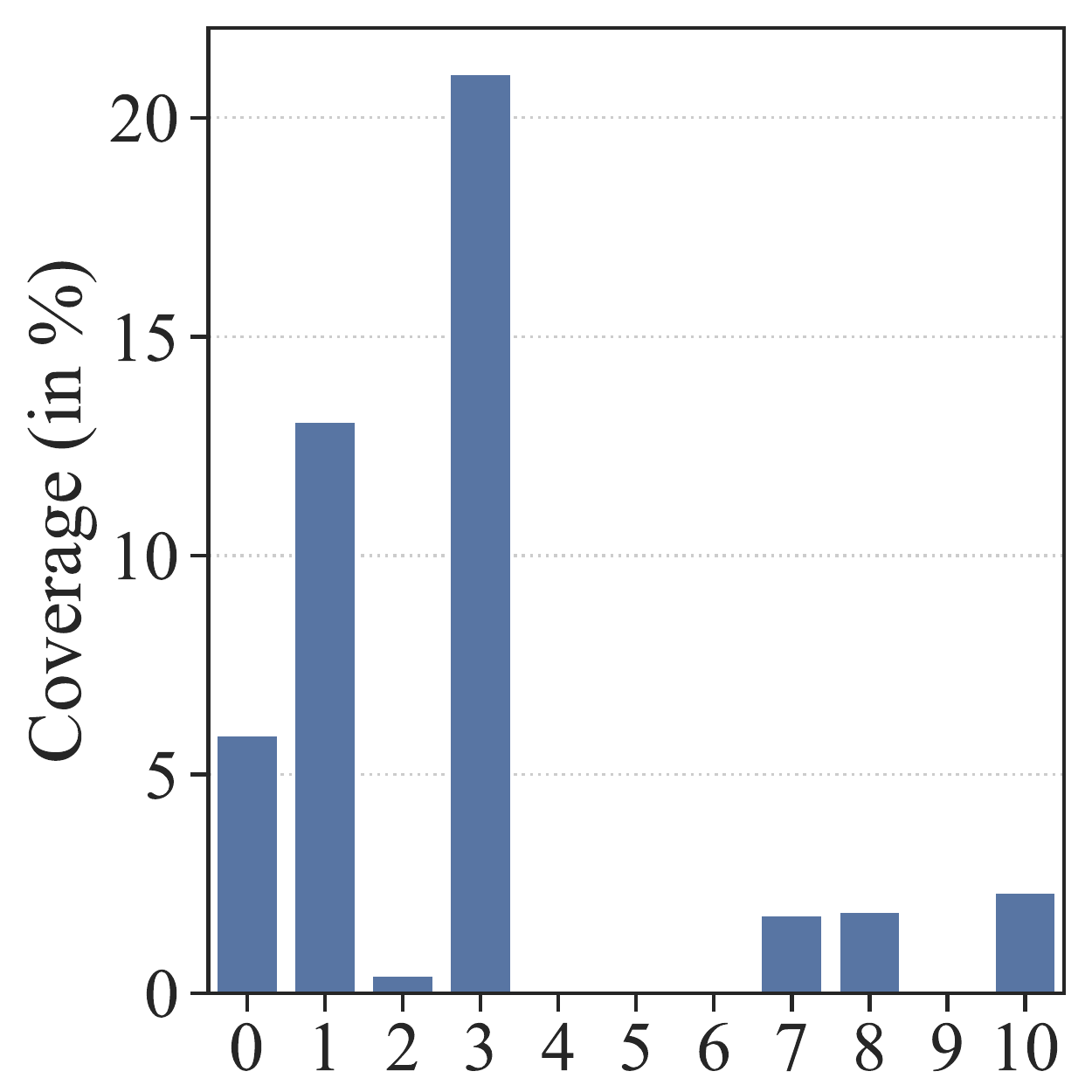}
		\label{fig:mesh_recall}
	}
	\hspace{-1.5ex}
	\subfloat[Performance.]{
		\includegraphics[width=0.31\linewidth]{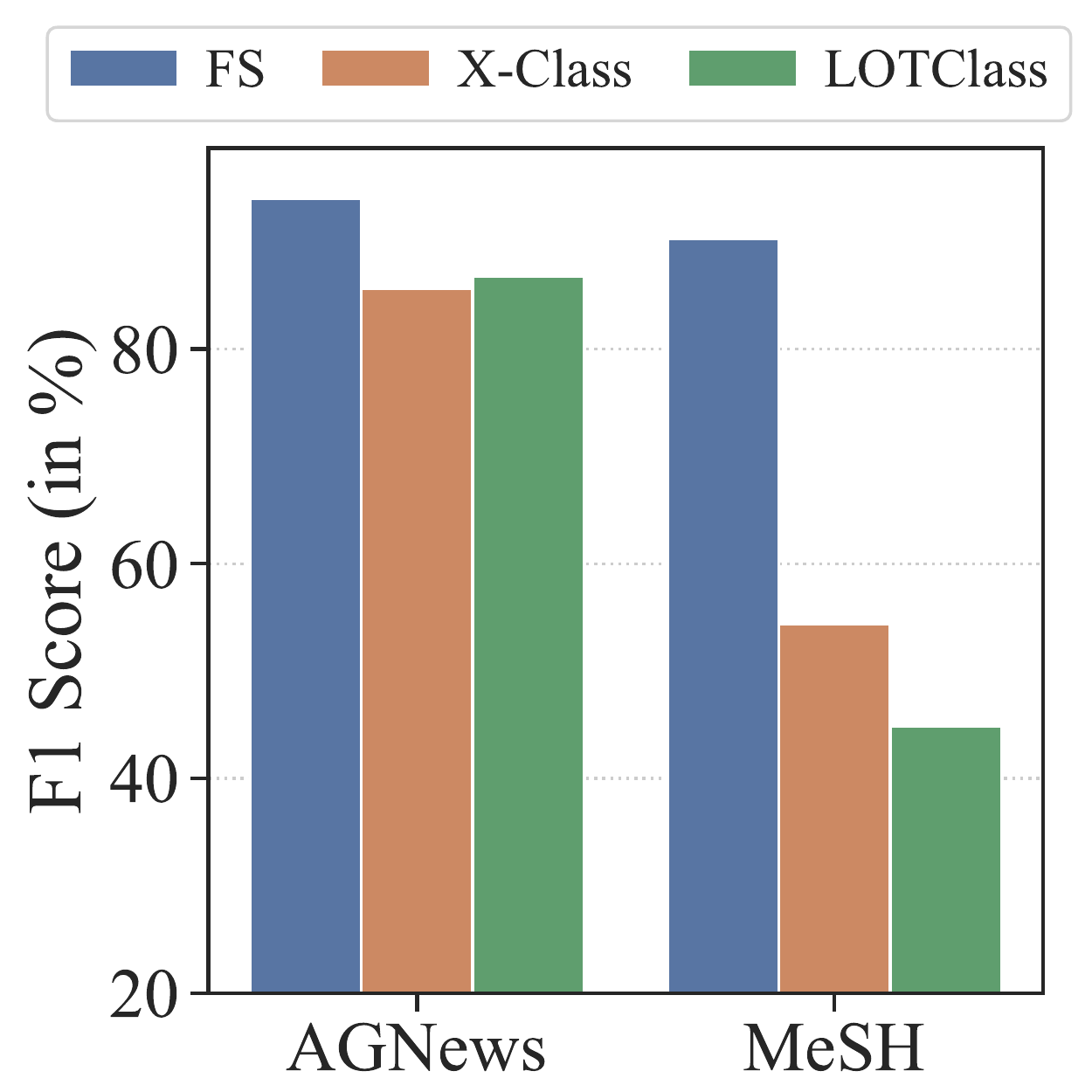}
		\label{fig:f1_baselines}
	} 
	\caption{Pilot Studies. FS means \emph{fully-supervised} model.}
	\vspace{-0.5ex}
\label{fig:pilot}
\end{figure}

While existing weakly supervised methods~\cite{lotclass,xclass} achieve competitive performance on general-domain datasets, applying them directly to scientific datasets often causes performance degradation. 
To illustrate this, we use AGNews~\cite{Zhang2015CharacterlevelCN} as the general-domain dataset and MeSH~\cite{cohan2020specter} as the scientific dataset. 
The average \emph{precision} (i.e. the portion of correctly matched examples) and \emph{coverage} (i.e. the portion of examples that can be matched by label names)
for label names are shown in Figure~\ref{fig:precision_recall}. 
We observe that for the scientific domain, the precision and coverage decline by 6\% and 37\% respectively. 
Moreover, the results of per-class coverage (presented in Figure~\ref{fig:mesh_recall}) indicate that the label distribution is more \emph{imbalanced} for scientific data. For MeSH, there are 4 out of 11 classes where the label name cannot match any examples from the unlabeled corpus.  

These two issues prevent the previous weakly-supervised models \cite{lotclass,xclass} from performing well. As shown in Figure~\ref{fig:f1_baselines}, gaps to the performance of the fully-supervised model are much larger for scientific datasets (36\%) than general-domain datasets (8\%), which indicates that these advanced techniques cannot resolve the unique challenges that exist in the scientific domain. 

\vspace{-1ex}
\section{Method}
From the analysis in the above section, we conclude that it is necessary to propose  techniques beyond hard matching to better harvest the semantic label name information. 
Towards this goal, we present our framework {\ours} in Figure~\ref{fig:framework}, a multi-stage training scheme based on dense retrieval, to perform document classification using label names only. 
The three stages are detailed below.

\vspace{-1ex}
\subsection{Stage-I: Dense Retrieval with Label Names}  
\label{sec:stage1}
Directly using the label-indicative keywords to extract documents is sub-optimal for scientific documents, due to their limited coverage and inferior ability to capture the class-related semantics. 
Motivated by this, we propose to leverage \emph{dense retrieval} (DR)~\cite{dpr} to effectively retrieve the most relevant documents. 
Specifically, DR represents the  input information (“query”) $q$ and  target corpus (“document”) $d$ in the continuous embedding space as $g(q; \phi), g(d; \phi)$ respectively, where $g(\cdot;\phi)$ is the dense retrieval model with $\phi$ being the parameter of $g$.
Then, DR matches queries and documents via approximate nearest neighbor (ANN)  using the relevance score $r(q, d;\phi) = \langle g(q;\phi), g(d;\phi) \rangle$, where $\langle\cdot,\cdot\rangle$ is the dot product.
Next, we introduce the approach to train the DR model as well as leverage DR to extract documents from the corpus $\cX$. 

\noindent $\square$ \textbf{Task-adaptive DR Model Pretraining.} 
To pretrain a DR model $g(\cdot;\phi)$ in an unsupervised fashion, we continuously pretrain the language model on the corpus $\cX$, using the contrastive learning widely adopted in recent research~\cite{izacard2022unsupervised,gao2022unsupervised,yu2022coco,zhang2023pre}.  
Specifically, for each document $d_i\in\cX$, we first split each document into multiple sentences and \emph{randomly} sample two sentences $d_{i,1}, d_{i,2}$ as the positive pair. The training objective of contrastive learning for $d_i$ is 
\begin{equation}
\setlength{\abovedisplayskip}{5pt}
\setlength{\belowdisplayskip}{5pt}
\ell_{\text{CL}}=-\log \frac{\exp{\left(\tau \cdot \langle g(d_{i,1};\phi),  g(d_{i,2};\phi) \rangle \right)}}{\sum_{j=1,2}\sum_{d^- \in \cD_i^{-}} {\exp}\left(\tau \cdot \langle g(d_{i,j};\phi), g(d^-_i;\phi) \rangle \right)},
\label{eq:cl}
\end{equation}
where $d^-_i\in\cD_i^{-}$ are the in-batch negatives, and $\tau=0.01$ is the parameter for temperature. Contrastive pretraining improves both the alignment and  uniformity for embeddings of sequences~\cite{wang2020understanding, zhu2022structure, kan2022fbnetgen, xu2022counterfactual,yang2022pre}, which can better support the  retrieval task. \\
\noindent$\square$ \textbf{Document Retrieval using Label Names.}  
With the DR model, we aim to extract an initial set of labeled data for each class by feeding the label names (as queries) to the DR model. The initial retrieved document set $\cD_{i}$ for the $i$-th class can be written as 
\begin{equation}
\setlength{\abovedisplayskip}{5pt}
\setlength{\belowdisplayskip}{5pt}
\cD_{i}=\operatorname{Top-}k_{d \in \cX}^{\mathrm{ANN}} r(w_i, d ; \phi),
\label{eq:knn}
\end{equation}
where $k$ is the number of retrieved examples, and the label of the retrieved document is determined by the category of the label name. 
In this way, we get rid of the challenge brought by those infrequent label names and provide a flexible way to encode the label-related semantics. All retrieved examples $\cD=\cup_{i=1}^{C}\cD_{i}$ are then used for classification, which will be discussed in the following part.
\vspace{-0.2ex}
\begin{table*}[!t]
    \centering
     \vspace{-1ex}
     \caption{Label Name Information. Label names in \blue{blue} never appear in the corpus.}
     \vspace{-1ex}
    \renewcommand\arraystretch{0.9}
\resizebox{\linewidth}{!}{
\begin{tabular}{cl}
\toprule
    \bf Dataset & \bf Label Names   \\\midrule
    \multirow{2}{*}{\bf MeSH} & 	 Cardiovascular diseases,  Chronic kidney disease, HIV/AIDS, Diabetes (mellitus), \blue{Chronic respiratory diseases}, \blue{Digestive diseases}, \blue{Hepatitis A/B/C/E},  \\
    &    Mental disorders,  Musculoskeletal disorders, \blue{Neoplasms (cancer)}, Neurological disorders	 \\ \midrule
    \multirow{2}{*}{\bf arXiv-Math} & Numerical Analysis,  Algebraic Geometry, Functional spaces, Number Theory,  Complex Variable
, Differential Geometry, Combinatorics, Operator Algebra,  \\
&  Representation Theory, \blue{Statistics Theory}, \blue{Topological Geometry}, Rings and Algebra, Probability
, Dynamical System, \blue{Optimization and Control}, Logic \\\midrule
    \multirow{4}{*}{\bf arXiv-CS} & Database, Computation and Language, Information Theory, Computational Geometry,  \blue{Cryptography and Security}, \blue{System and Control}, Game Theory,   \\
& \blue{Data Structures and Algorithm}, Human-Computer Interaction, Machine Learning, Information Retrieval, Programming Languages,  Software Engineering, \\
&  \blue{Networking and Internet Architecture}, Artificial Intelligence, Social and Information Networks,  Distributed, Parallel, and Cluster Computing, Robotics,  \\ & Computer Vision and Pattern Recognition, \blue{Logic in Computer Science}\\
    \bottomrule
\end{tabular}}
\label{tab:label_name}
\end{table*}

\noindent $\square$ \textbf{Training Classifiers with Retrieved Text.} With the retrieved document set $\cD$, one can simply finetune a classifier $f(\cdot;\theta)$ with the standard cross-entropy loss:
\begin{equation}
\setlength{\abovedisplayskip}{5pt}
\setlength{\belowdisplayskip}{5pt}
\min_{\theta}~\mathbb{E}_{(x_{i}, y_{i}) \in \cD} ~~\ell_{\text{CE}}\left(f(x_{i}; \theta), y_{i}\right).
\label{eq:cross_entropy_loss}
\end{equation}
The fine-tuned model is used for target classification tasks.
\vspace{-1ex}
\subsection{Stage-II: Expand Label Names with Local and Global Information} 
\label{sec:stage2}
One drawback of the above stage is that the label names are often too abstract to fully represent the semantics information for classes.
As such, the retrieved documents still contain label noise, which hurts the downstream performance. 
To tackle this, we propose to automatically extract class-related keywords to expand the label name, by using both local information from the retrieved documents and global information from the general pretrained models. 

\noindent $\square$ \textbf{Local Information for Keyword Extraction.} To identify the class-related keywords, we assume terms that appear frequently within documents from a specific class while infrequently for other classes are more likely to be class-indicative words for that class. 
Inspired by TF-IDF ~\cite{grootendorst2022bertopic}, we measure the indicativeness of word $w$ for class $c$ from the retrieved document $\cD$ as
\begin{equation}
\setlength{\abovedisplayskip}{5pt}
\setlength{\belowdisplayskip}{5pt}
L_{w, c}=\operatorname{tf}^{\alpha}_{w, c} \cdot \log \left(1+{A}/{\operatorname{tf}_w}\right) \cdot \operatorname{cnt}_{w, c}.
\label{eq:c_tf_idf}
\end{equation}
Here $\operatorname{tf}_{w, c}$, $\operatorname{cnt}_{w, c}$ stands for the frequency and occurrence time of word $w$ within documents from class $c$ and $\operatorname{tf}_w$, is the 
frequency of $w$ in corpus, $A$ is the average number of
words per class. 
In this way, words appear commonly in the class-related documents (higher $\operatorname{tf}_{w, c}$) 
while being less generic (lower $\operatorname{tf}_w$) will receive higher score. 
For each class, we extract $m$ words with the highest score as the candidate set $\cC$.\footnote{We omit words that already appeared in label names during the expansion (stage-II).}  

\noindent $\square$ \textbf{Global Information for Keyword Semantics.} The above step only considers the word occurrence in the local corpus, without modeling the semantic information. An ideal keyword, however, should also have a closer meaning to the label name. 
Motivated by this, we  leverage the PTLM to transfer the \emph{global} knowledge from pretraining corpora and encode the contextual information for each word. 
We calculate the embeddings of both label names and candidate words by averaging the output of all tokens from the last layer of PTLM $h(\cdot;\psi)$. 
For word $w\in \cC$ from the candidate set of class $c$, the global score is calculated between $w$ and the label name $w_c$ using the embedding similarity as  
\begin{equation}
\setlength{\abovedisplayskip}{5pt}
\setlength{\belowdisplayskip}{5pt}
G_{w, c}=  \cos\left( h(w; \psi), h(w_c; \psi) \right).
\label{eq:global_score}
\end{equation}
\noindent $\square$ \textbf{Ensemble Reranking.} 
To effectively combine the local and global information, we sort  candidate words $w\in\cC_i$ for $i$-th class using the score $L_{w,c}, G_{w,c}$, respectively. 
Then, each word $w$ will have two ranks as $\operatorname{rank}_{L,c}(w)$ and $\operatorname{rank}_{G,c}(w)$. We rerank the words using the
ensemble score based on Reciprocal Rank Fusion (RRF)~\cite{cormack2009reciprocal}:
\begin{equation}
\setlength{\abovedisplayskip}{5pt}
\setlength{\belowdisplayskip}{5pt}
\operatorname{score}_{w, c}= 1/\operatorname{rank}_{G,c}(w) + 1/\operatorname{rank}_{L,c}(w).
\label{eq:ensemble_score}
\end{equation}
For each class, we add one word with the highest score to expand the label name. For expansion, we simply concatenate the previous label name and the newly identified word for enrichment~\cite{cao2008selecting}.

\noindent $\square$ \textbf{Iterative Label Name Expansion.}  The above process can be conducted multiple times. In each iteration, we first use local and global scores to detect the expanded words using Eq.~\eqref{eq:c_tf_idf}--\eqref{eq:ensemble_score} and enrich the label names. 
Then, we use the expanded label names as queries to update the retrieved documents $\cD$ with Eq.~\eqref{eq:knn} as we expect the quality of $\cD$ will improve by incorporating additional class-indicative words. 
With the updated $\cD$, more relevant words can be extracted  to enrich the class information. The above iteration is repeated 5 times, and the retrieved documents after the final iteration can be used to train another classifier using Eq.~\eqref{eq:cross_entropy_loss}.
\vspace{-1.5ex}

\subsection{Stage-III: Refine Classifier with Self-training} 
\label{sec:stage3}
The pseudo-labeled samples in Stage-II are only from the top retrieved documents with the expanded label names.
To generalize its current knowledge to the whole unlabeled corpus, \emph{self-training} is adopted to bootstrap the model on the entire unlabeled corpus~\cite{lotclass,yu-etal-2022-actune} as
\begin{equation}
\setlength{\abovedisplayskip}{5pt}
\setlength{\belowdisplayskip}{5pt}
\min_{\theta} ~\mathbb{E}_{(x, \tilde{y}) \in {\cX}} ~~~\mathbbm{1}\left\{[f({x};  \theta)]_{\tilde{y}}>\gamma \right\}  \times 
 \ell\textsubscript{CE} \left( f(x; {\theta}), \tilde{y} \right),
\label{eq:st}
\end{equation}
where $\tilde{y}=\argmax f(x; {\theta})$ is the hard pseudo label, $\gamma$ is the confidence threshold.
With self-training, the model is refined by its high-confident predictions to improve generalization ability.
Stage-III stops 
when less than 1\% of samples change their labels, and the final model can be used to classify any document.

\vspace{-1ex}
\section{Experiments}
\begin{table}
    \centering
     \caption{Dataset statistics.}
     \vspace{-2ex}
    \renewcommand\arraystretch{0.9}
\resizebox{\linewidth}{!}{
\begin{tabular}{ccccccc}
\toprule
    \bf Dataset & \bf Domain &\bf \# Train &\bf \# Test  &\bf \# Class& \bf \# OOV  &\bf Avg. Len.   \\\midrule
    MeSH & BioMedical  & 16.3k & 3.5k & 11 & 4 (36\%)  &  254.3 \\
    arXiv-Math & Mathematics & 62.5k & 6.3k & 16  & 3 (19\%) & 214.4 \\ 
    arXiv-CS & Computer Science  & 75.7k & 5.1k & 20 & 5 (25\%) & 188.2 \\ 
    \bottomrule
\end{tabular}}
\label{tab:dataset}
\end{table}

\vspace{-0.1ex}
\subsection{Experiment Setups}
\noindent $\square$ \textbf{Datasets.} We conduct experiments on three datasets from multiple domains including MeSH~\cite{cohan2020specter}, arXiv-CS~\cite{clement2019use}, arXiv-Math~\cite{clement2019use}. {The  statistics and label names for each dataset are shown in Table~\ref{tab:label_name} and~\ref{tab:dataset}}. For arXiv-CS and arXiv-Math, we select papers from years 2017-2020 as the training set, 2021-2022 as the test set, and  use the topic from \emph{the main category} as the label.

\noindent $\square$ \textbf{Baselines.} We compare {\ours} with these baselines: 
(1) \underline{\textbf{IR}}~\cite{tfidf} leverages TF-IDF to assign labels for documents. 
(2) \underline{\textbf{Dataless}}~\cite{dataless} uses Wikipedia to embed labels and documents. Each document is classified
to the label with the highest similarity.
(3) \underline{\textbf{SentenceBERT}} \cite{sentencebert} is trained on NLI data to embed labels and documents for classification.
(4) \underline{\textbf{LOTClass}} \cite{lotclass} and (5) \underline{\textbf{X-Class}} \cite{xclass} are two methods that use PTLMs for label-name-only text classification by using masked language modeling or pretrained representations.
(6) \underline{\textbf{FastClass}} \cite{xia2022fastclass} uses SentenceBERT to extract initial labeled examples, then selects an optimal subset for classifier training.

\noindent $\square$ \textbf{Implementations.}  
We use the pre-trained SciBERT~\cite{beltagy2019scibert} as the backbone for  {\ours} and baselines.  
The retrieval model $g$ (Eq.~\eqref{eq:cl}) and PTLM $h$ (Eq.~\eqref{eq:global_score}) are initialized from SciBERT, and $g$ is pretrained on the  corpus $\cX$ for 5 epochs. Note that to avoid information leakage, \emph{only the training set is used for pretraining}.
The maximum length is set to 512. 
For Stage-I and II, we finetune $f(\cdot;\theta)$ for 5 epochs with Adam as the optimizer and set the batch size and learning rate to 32 and 2e-5. 
Other hyperparameters include $\tau$ in Eq.~\eqref{eq:cl}, $k$ for ANN in Eq.~(\ref{eq:knn}), $\gamma$ in Eq.~(\ref{eq:st}), $m$ in Sec.~\ref{sec:stage2}.
We set $\tau=0.01,m=100,k=100,\gamma=0.8$ without tuning. 
\begin{table}[!t]
\centering
\caption{Performance on three datasets.  
\textbf{Bold} and \blue{blue} indicate the best and second-best results for each dataset. \emph{Macro-F1} is the main metric as the label distribution is imbalanced.
}
\renewcommand\arraystretch{0.9}
  \resizebox{1.01\linewidth}{!}{%
\begin{tabular}{lccc ccc }
\toprule
\multirow{2.5}{*}{\bf Method} & \multicolumn{2}{c}{\bf MeSH}    & \multicolumn{2}{c}{\bf arXiv-Math}    & \multicolumn{2}{c}{\bf arXiv-CS}         \\ \cmidrule(lr){2-3} \cmidrule(lr){4-5} \cmidrule(lr){6-7}  
& Mi-F1 & Ma-F1         & Mi-F1 & Ma-F1        & Mi-F1 & Ma-F1        \\ \midrule
 Fully Supervised &  90.5\scriptsize±0.3 &	90.3\scriptsize±0.2	 &80.6\scriptsize±0.4 &	79.1\scriptsize±0.3 &	83.0\scriptsize±0.2	 &78.2\scriptsize±0.4 \\ 
  \midrule
 IR~\cite{tfidf}  &  40.6 & 37.6 & 27.8 & 22.9 & 24.5 & 22.8 \\
Dataless~\cite{dataless} & 36.1 & 26.8 & 18.9	 & 13.4	 & 20.5	 &18.2  \\ 
SentenceBERT~\cite{sentencebert} &68.6	 &66.0 &	48.9 &	41.1 &	50.7 & 47.7 \\ 
LOTClass~\cite{lotclass} &   57.9\scriptsize±1.7&	44.9\scriptsize±1.6	&43.8\scriptsize±2.0&	35.2\scriptsize±1.5&	51.5\scriptsize±1.4&	47.1\scriptsize±1.8    \\ 
X-Class~\cite{xclass}  &  55.2\scriptsize±1.4	&54.4\scriptsize±1.8	&46.5\scriptsize±1.4&	39.1\scriptsize±1.4	&\blue{60.6\scriptsize±1.2} &\blue{51.6\scriptsize±1.3} \\ 
FastClass~\cite{xia2022fastclass} & \blue{78.5\scriptsize±1.3}	&\blue{78.1\scriptsize±1.1}&	\blue{53.5\scriptsize±1.3}	&\blue{44.5\scriptsize±1.2}	&59.8\scriptsize±0.8&	50.5\scriptsize±0.9  \\ 
\midrule
\rowcolor{gray!20} {\ours}                                       & \textbf{82.0\scriptsize±0.4}  & \textbf{81.9\scriptsize±0.4}  & \textbf{58.0\scriptsize±0.8}  & \textbf{51.9\scriptsize±0.7}      & \textbf{65.6\scriptsize±0.8}  & \textbf{58.9\scriptsize±0.6}         \\ 
\rowcolor{gray!20} Gain $\Delta$  &  3.5 \scriptsize(4.4$\%$) &  3.8 \scriptsize(4.9$\%$) &  4.5 \scriptsize(8.4$\%$) & 
 7.4 \scriptsize(16.6$\%$) &  5.0 \scriptsize(8.2$\%$) &   7.3 \scriptsize(14.1$\%$)  \\ \midrule
 {\ours (Stage-I)}                                       & 76.6\scriptsize±1.0	 & 75.6\scriptsize±0.8	 & 56.4\scriptsize±1.4	 & 49.8\scriptsize±0.9	 & 61.8\scriptsize±1.1	 & 54.7\scriptsize±1.2
     \\ 
{\ours (Stage-II)}                                       & 79.9\scriptsize±0.6	 & 80.2\scriptsize±0.7	 & 57.1\scriptsize±1.1	 & 51.0\scriptsize±1.0	 & 64.6\scriptsize±1.0	 & 58.1\scriptsize±0.6         \\ 
 \bottomrule
\end{tabular}
}
\label{tab:Main}
\vspace{-0.1ex}
\end{table}

\vspace{-0.5ex}
\subsection{Experiment Results}
\noindent $\square$ \textbf{Main Experiments.} 
We report both Macro-F1 and Micro-F1 scores for {\ours} and baselines in Table~\ref{tab:Main}.
The mean and variance over 5 runs are calculated when  fine-tuning is used.
We observe that {\ours} consistently achieves the best performance on three datasets, with an average gain of 11.9\%. 
In contrast, X-Class and LOTClass, which achieve strong results on general-domain tasks, fail to perform well on the scientific domain, as they cannot handle the challenges mentioned in Sec.~\ref{sec:challenge}. 
Moreover, traditional baselines, such as IR, and Dataless,  are inferior to other methods using PTLMs, indicating their limited ability for modeling scientific text. Although SentenceBERT and FastClass use extra labeled data for embedding learning, distribution shifts exist between the labeled data and scientific corpus. They also fail to expand the label names for enriching representations, leading to sub-optimal performance.

\noindent $\square$  \textbf{Effect of Multi-stage Training.} The bottom two rows in Table~\ref{tab:Main} show the performance of {\ours} after Stage-I and II, which justifies that all three stages contribute to the final performance. Moreover, {\ours} outperforms all baselines even without self-training (Stage-III), indicating that it can retrieve a small set of high-quality data to support downstream tasks sufficiently.
\begin{figure}[t]
    \centering
    \vspace{-0.5ex}
    \subfloat[Study of Retrievers]{
        \includegraphics[width=0.325\linewidth]{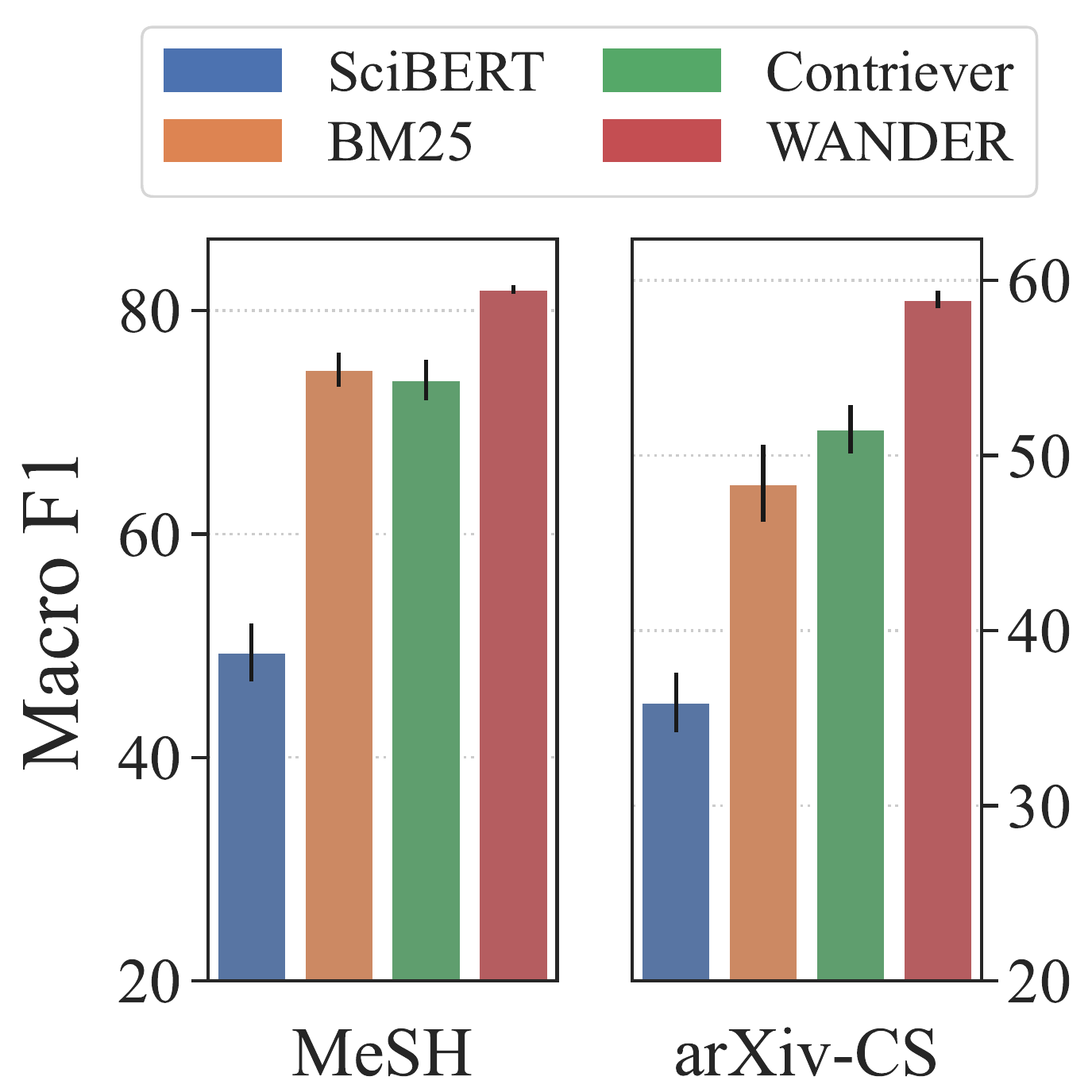}
        \label{fig:dr_models}
    } 
        \hspace{-1.5ex}
    \subfloat[Study on $k$]{
        \includegraphics[width=0.325\linewidth]{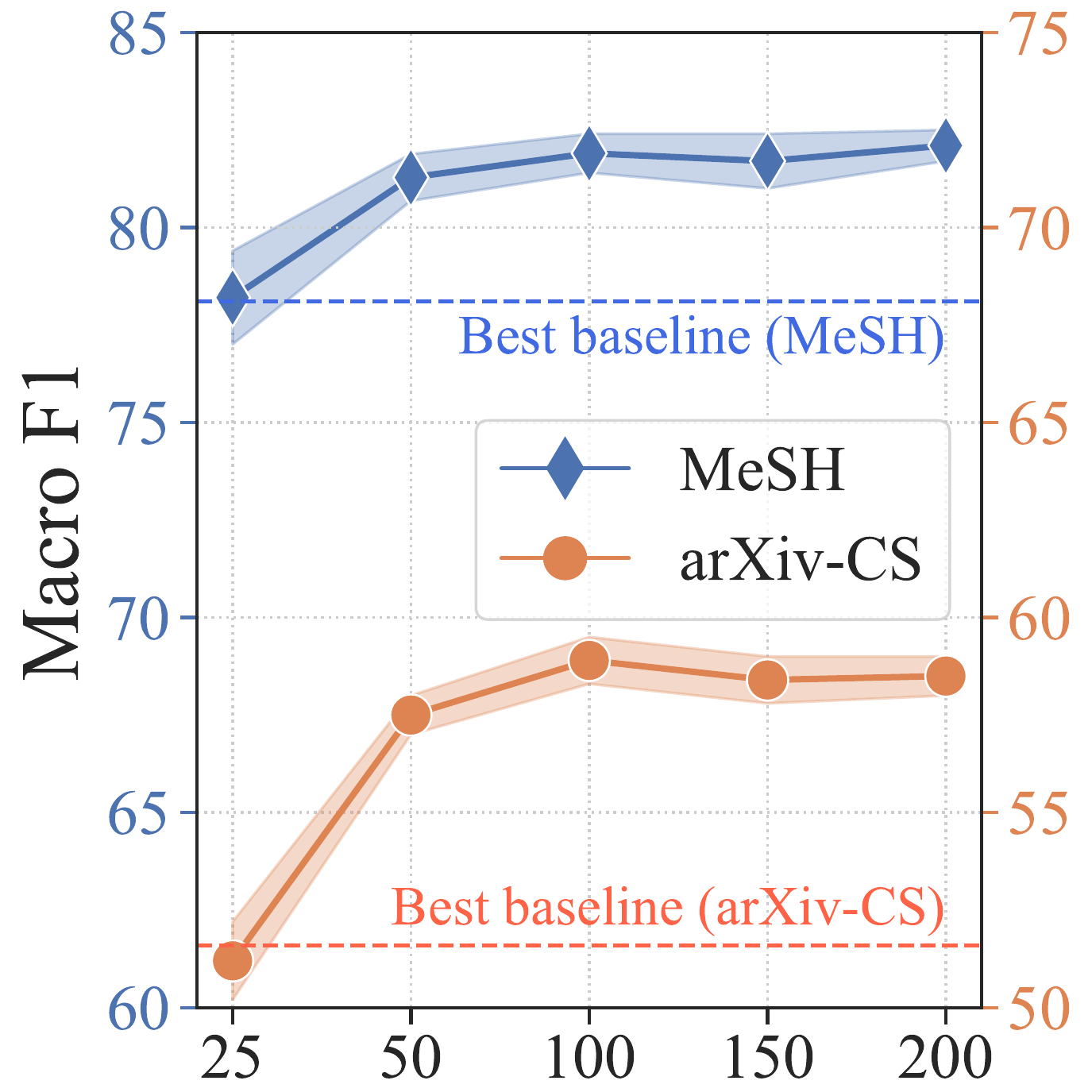}
        \label{fig:para_k}
    }
    \hspace{-1.5ex}
    \subfloat[Study on $\gamma$]{
        \includegraphics[width=0.325\linewidth]{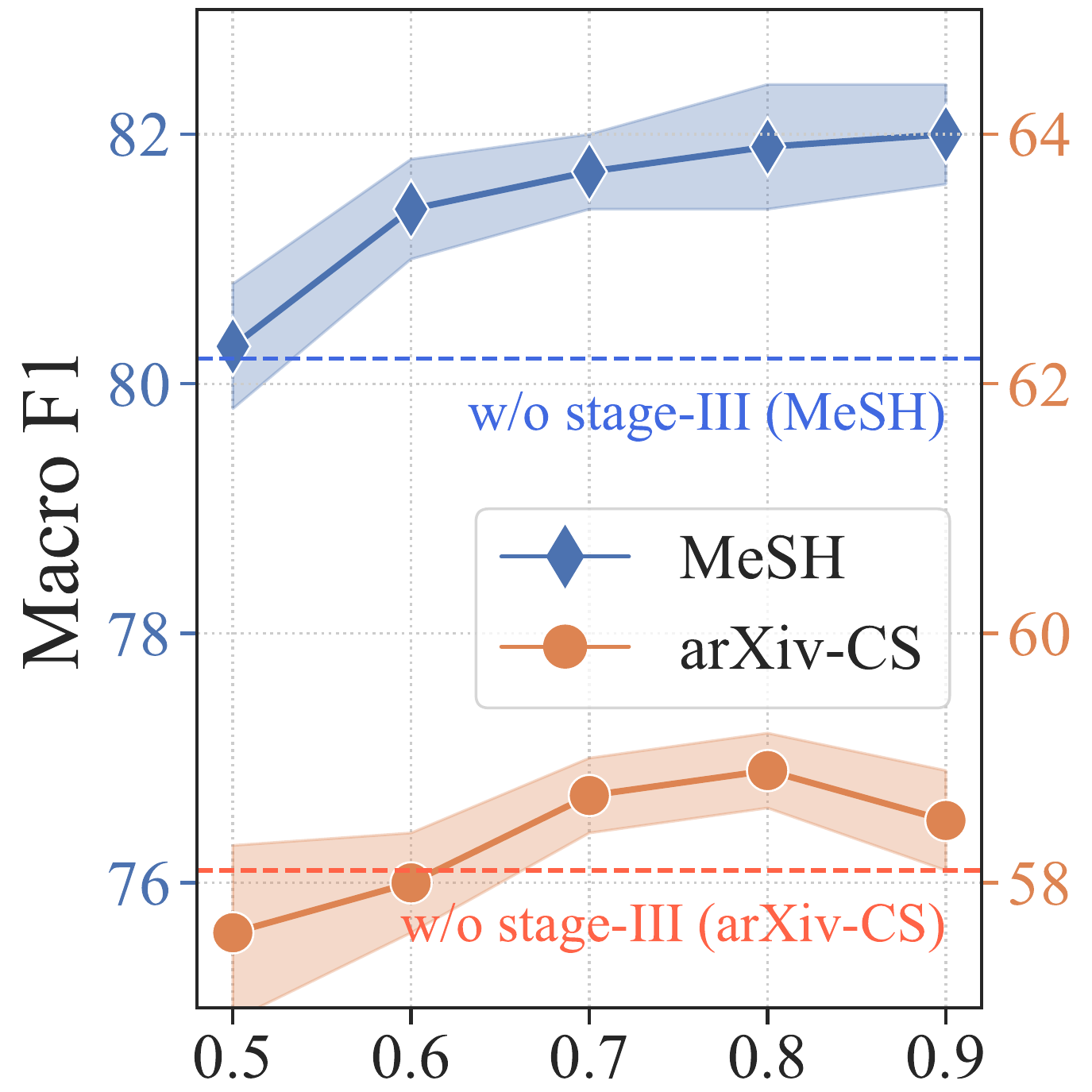}
        \label{fig:para_gamma}
    }
    \vspace{-0.5ex}
    \caption{Studies of Different Retrieval Models and Hyperparameters (Best View in Colors).}
    \vspace{-1ex}
\label{fig:dr_model}
\end{figure}

\begin{figure}[t]
    \centering
    \vspace{-1.5ex}
    \subfloat[Pseudo Label Accuracy]{
        \includegraphics[width=0.47\linewidth]{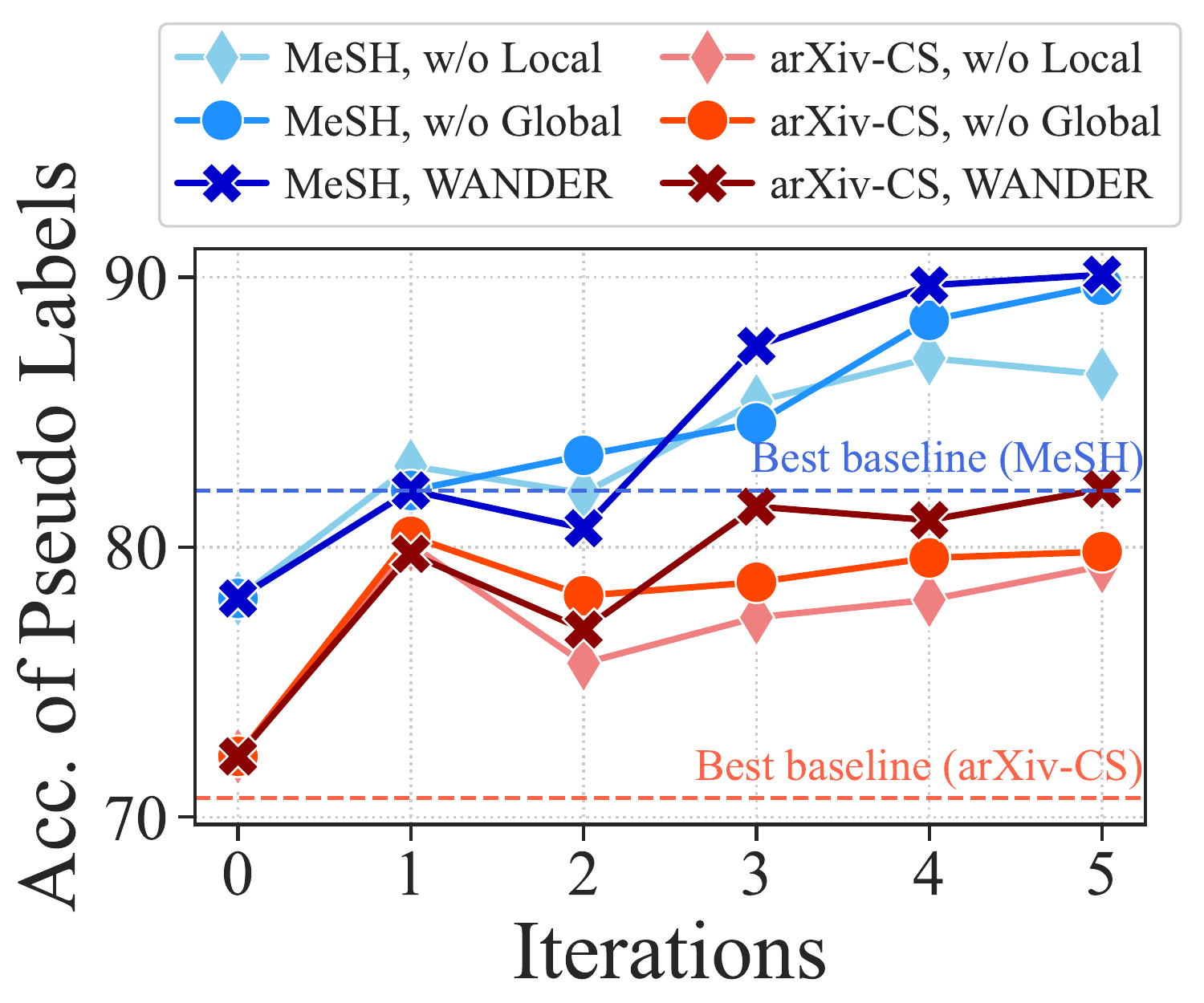}
        \label{fig:local_global_acc}
    } 
         \hspace{-1ex}
    \subfloat[Macro-F1 Score]{
        \includegraphics[width=0.47\linewidth]{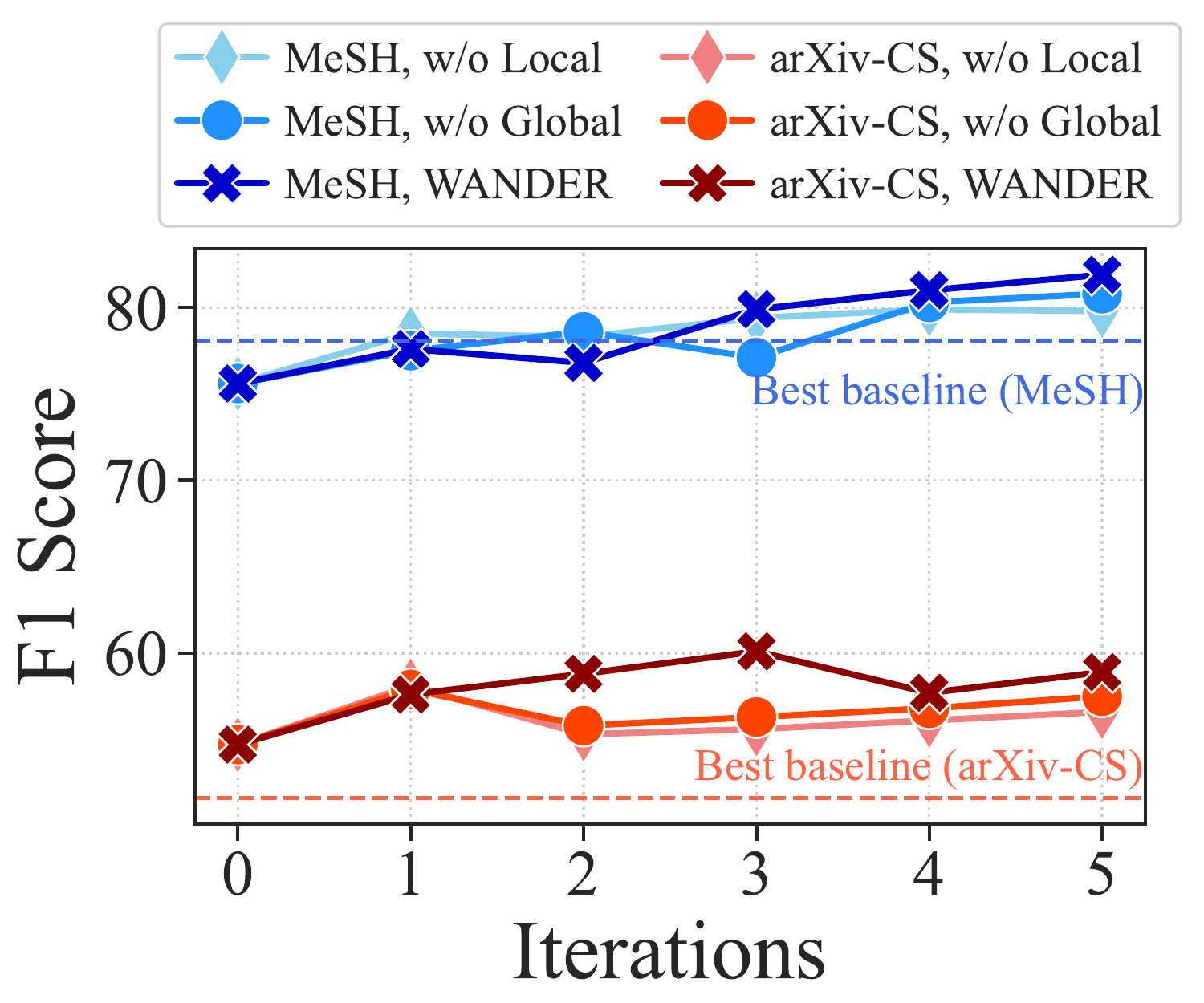}
        \label{fig:local_global_f1}
    }
    \caption{Study on Effects of Local and Global Information.}
    \vspace{-1ex}
\label{fig:local_global}
\end{figure}

\subsection{Ablation and Hyperparameter Studies}
\label{sec:ablation}
\noindent $\square$  \textbf{Study of DR Models.} 
To illustrate the effect of task-adaptive contrastive learning (TAPT) for DR model pretraining, we substitute $g(\cdot)$ with other models including \emph{BM25}~\cite{bm25}, \emph{SciBERT}~\cite{beltagy2019scibert} without TAPT, the strong unsupervised DR model \emph{Contriever}~\cite{izacard2022unsupervised}, and compare the performance in Figure~\ref{fig:dr_models}. 
Overall, our model achieves the best performance, which justifies the need for TAPT as it effectively reduces the distribution shifts and also produces better embeddings. Instead, using sparse retrieval model (BM25) yields undesirable performance as it cannot understand label names well.

\noindent $\square$  \textbf{Effect of Hyperparameters.} 
We study the effect of $k$ and $\gamma$ in {\ours} on MeSH and arXiv-CS, as shown in Figure~\ref{fig:para_k} and~\ref{fig:para_gamma}. We observe that the performance first increases with larger $k$ as the model benefits from more retrieved examples. When $k$ reaches 100, the performance remains stable, as too many retrieved examples introduce label noise and diminish the performance gain.
We also run experiments with different thresholds $\gamma$. The result indicates that the model performance is insensitive to $\gamma$, and the self-training component leads to performance gain in most studied regions.

\noindent $\square$  \textbf{Effect of Local and Global Information.} Figure~\ref{fig:local_global} illustrates the performance of {\ours} and its variants over 5 expansion iterations. Overall, we observe that removing local or global information hurts the performance, since these two modules provide complementary information. Combining these two terms together results in better pseudo labels and improves downstream performance.

\subsection{Case Studies}
\begin{table}[!t]
    \centering
     \caption{Case Study on expanded keywords for three tasks.}
    \renewcommand\arraystretch{0.9}
\resizebox{1.01\linewidth}{!}{
\begin{tabular}{ccc}
\toprule
    \bf Dataset & \bf Class &\bf  Expanded Keyword  \\\midrule
    MeSH & Diabetes  & insulin, glucose, diabetic, metformin, glycemic\\
    MeSH & Mental Disorders  & depression, anxiety, antidepressant, schizophrenia, mood \\ 
    MeSH & Neoplasms  & tumor, carcinoma, cell, tumour, chemotherapy\\\midrule
    arXiv-Math & Combinatorics & graph, combinatorial, vertex, edge,  bipartite \\
    arXiv-Math & Number theory &  prime, integer, modulo, odd, divisor \\
    arXiv-Math & Statistics theory & estimation, sample, regression, treatment, inference \\ \midrule
    arXiv-CS & Information theory & entropy, channel, shannon, capacity, decoder \\
    arXiv-CS & Machine Learning & classifier, classification, boosting, ensemble, tree \\
    arXiv-CS & Game Theory  & player, equilibrium, nash, payoff, strategy \\
    \bottomrule
\end{tabular}}
\label{tab:case}
\end{table}
We present a case study in Table~\ref{tab:case}  to showcase that {\ours} is able to discover class-related keywords to expand label names. Take \emph{diabetes} as an example, it is often related to high \emph{glucose} level and \emph{glycemic} index. Besides, \emph{insulin} and \emph{metformin} are used as treatments for diabetes. Moreover, take \emph{machine learning} as another example, it is applied to \emph{classifiation} tasks. \emph{Boosting}, \emph{ensemble}, \emph{tree} are all techniques to tackle machine learning problems. These all indicate that {\ours} can enrich the semantics of label names.

\section{Conclusion}
We propose {\ours}, a multi-stage training framework for weakly-supervised scientific document classification with label name only.
We leverage \emph{dense retrieval} to go beyond hard matching and harness the semantics of label names. In addition, we propose a label name expansion module to enrich its representations, and use self-training to improve the model's generalization ability. 
Experiments on three datasets demonstrate that {\ours} outperforms the baselines by 11.9\% on average.
For future works, we plan to extend {\ours} to other scenarios such as multi-label classification. 
\section*{Acknowledgements}
We thank the anonymous reviewers and area chairs for the valuable feedbacks. 
This research was partially supported by the internal funds and GPU servers provided by the Computer Science Department of Emory University. 
JH was supported by NSF grants IIS-1838200 and IIS-2145411. 
\balance
\bibliographystyle{ACM-Reference-Format}
\bibliography{sample-base}


\begin{thebibliography}{40}


\ifx \showCODEN    \undefined \def \showCODEN     #1{\unskip}     \fi
\ifx \showDOI      \undefined \def \showDOI       #1{#1}\fi
\ifx \showISBNx    \undefined \def \showISBNx     #1{\unskip}     \fi
\ifx \showISBNxiii \undefined \def \showISBNxiii  #1{\unskip}     \fi
\ifx \showISSN     \undefined \def \showISSN      #1{\unskip}     \fi
\ifx \showLCCN     \undefined \def \showLCCN      #1{\unskip}     \fi
\ifx \shownote     \undefined \def \shownote      #1{#1}          \fi
\ifx \showarticletitle \undefined \def \showarticletitle #1{#1}   \fi
\ifx \showURL      \undefined \def \showURL       {\relax}        \fi
\providecommand\bibfield[2]{#2}
\providecommand\bibinfo[2]{#2}
\providecommand\natexlab[1]{#1}
\providecommand\showeprint[2][]{arXiv:#2}

\bibitem[Beltagy et~al\mbox{.}(2019)]%
        {beltagy2019scibert}
\bibfield{author}{\bibinfo{person}{Iz Beltagy}, \bibinfo{person}{Kyle Lo},
  {and} \bibinfo{person}{Arman Cohan}.} \bibinfo{year}{2019}\natexlab{}.
\newblock \showarticletitle{SciBERT: A Pretrained Language Model for Scientific
  Text}. In \bibinfo{booktitle}{\emph{EMNLP-IJCNLP}}.
  \bibinfo{pages}{3615--3620}.
\newblock


\bibitem[Cao et~al\mbox{.}(2008)]%
        {cao2008selecting}
\bibfield{author}{\bibinfo{person}{Guihong Cao}, \bibinfo{person}{Jian-Yun
  Nie}, \bibinfo{person}{Jianfeng Gao}, {and} \bibinfo{person}{Stephen
  Robertson}.} \bibinfo{year}{2008}\natexlab{}.
\newblock \showarticletitle{Selecting good expansion terms for pseudo-relevance
  feedback}. In \bibinfo{booktitle}{\emph{SIGIR}}. \bibinfo{pages}{243--250}.
\newblock


\bibitem[Chang et~al\mbox{.}(2008)]%
        {dataless}
\bibfield{author}{\bibinfo{person}{Ming-Wei Chang}, \bibinfo{person}{Lev-Arie
  Ratinov}, \bibinfo{person}{Dan Roth}, {and} \bibinfo{person}{Vivek
  Srikumar}.} \bibinfo{year}{2008}\natexlab{}.
\newblock \showarticletitle{Importance of Semantic Representation: Dataless
  Classification.}. In \bibinfo{booktitle}{\emph{AAAI}}.
  \bibinfo{pages}{830--835}.
\newblock


\bibitem[Clement et~al\mbox{.}(2019)]%
        {clement2019use}
\bibfield{author}{\bibinfo{person}{Colin~B Clement}, \bibinfo{person}{Matthew
  Bierbaum}, \bibinfo{person}{Kevin~P O'Keeffe}, {and}
  \bibinfo{person}{Alexander~A Alemi}.} \bibinfo{year}{2019}\natexlab{}.
\newblock \showarticletitle{On the Use of ArXiv as a Dataset}.
\newblock \bibinfo{journal}{\emph{arXiv preprint arXiv:1905.00075}}
  (\bibinfo{year}{2019}).
\newblock


\bibitem[Cohan et~al\mbox{.}(2020)]%
        {cohan2020specter}
\bibfield{author}{\bibinfo{person}{Arman Cohan}, \bibinfo{person}{Sergey
  Feldman}, \bibinfo{person}{Iz Beltagy}, \bibinfo{person}{Doug Downey}, {and}
  \bibinfo{person}{Daniel~S Weld}.} \bibinfo{year}{2020}\natexlab{}.
\newblock \showarticletitle{SPECTER: Document-level Representation Learning
  using Citation-informed Transformers}. In \bibinfo{booktitle}{\emph{ACL}}.
  \bibinfo{pages}{2270--2282}.
\newblock


\bibitem[Cormack et~al\mbox{.}(2009)]%
        {cormack2009reciprocal}
\bibfield{author}{\bibinfo{person}{Gordon~V Cormack},
  \bibinfo{person}{Charles~LA Clarke}, {and} \bibinfo{person}{Stefan
  Buettcher}.} \bibinfo{year}{2009}\natexlab{}.
\newblock \showarticletitle{Reciprocal rank fusion outperforms condorcet and
  individual rank learning methods}. In \bibinfo{booktitle}{\emph{SIGIR}}.
  \bibinfo{pages}{758--759}.
\newblock


\bibitem[Cui et~al\mbox{.}(2022)]%
        {cui2022can}
\bibfield{author}{\bibinfo{person}{Hejie Cui}, \bibinfo{person}{Jiaying Lu},
  \bibinfo{person}{Yao Ge}, {and} \bibinfo{person}{Carl Yang}.}
  \bibinfo{year}{2022}\natexlab{}.
\newblock \showarticletitle{How Can Graph Neural Networks Help Document
  Retrieval: A Case Study on CORD19 with Concept Map Generation}. In
  \bibinfo{booktitle}{\emph{ECIR}}.
\newblock


\bibitem[Devlin et~al\mbox{.}(2019)]%
        {devlin2019bert}
\bibfield{author}{\bibinfo{person}{Jacob Devlin}, \bibinfo{person}{Ming-Wei
  Chang}, \bibinfo{person}{Kenton Lee}, {and} \bibinfo{person}{Kristina
  Toutanova}.} \bibinfo{year}{2019}\natexlab{}.
\newblock \showarticletitle{{BERT}: Pre-training of Deep Bidirectional
  Transformers for Language Understanding}. In
  \bibinfo{booktitle}{\emph{NAACL-HLT}}.
\newblock


\bibitem[Diaz et~al\mbox{.}(2016)]%
        {diaz2016query}
\bibfield{author}{\bibinfo{person}{Fernando Diaz}, \bibinfo{person}{Bhaskar
  Mitra}, {and} \bibinfo{person}{Nick Craswell}.}
  \bibinfo{year}{2016}\natexlab{}.
\newblock \showarticletitle{Query Expansion with Locally-Trained Word
  Embeddings}. In \bibinfo{booktitle}{\emph{ACL}}.
\newblock


\bibitem[Ganguly and Pudi(2017)]%
        {ganguly2017paper2vec}
\bibfield{author}{\bibinfo{person}{Soumyajit Ganguly} {and}
  \bibinfo{person}{Vikram Pudi}.} \bibinfo{year}{2017}\natexlab{}.
\newblock \showarticletitle{Paper2vec: Combining graph and text information for
  scientific paper representation}. In \bibinfo{booktitle}{\emph{ECIR}}.
  \bibinfo{pages}{383--395}.
\newblock


\bibitem[Gao and Callan(2022)]%
        {gao2022unsupervised}
\bibfield{author}{\bibinfo{person}{Luyu Gao} {and} \bibinfo{person}{Jamie
  Callan}.} \bibinfo{year}{2022}\natexlab{}.
\newblock \showarticletitle{Unsupervised Corpus Aware Language Model
  Pre-training for Dense Passage Retrieval}. In
  \bibinfo{booktitle}{\emph{ACL}}. \bibinfo{pages}{2843--2853}.
\newblock


\bibitem[Grootendorst(2022)]%
        {grootendorst2022bertopic}
\bibfield{author}{\bibinfo{person}{Maarten Grootendorst}.}
  \bibinfo{year}{2022}\natexlab{}.
\newblock \showarticletitle{BERTopic: Neural topic modeling with a class-based
  TF-IDF procedure}.
\newblock \bibinfo{journal}{\emph{arXiv preprint arXiv:2203.05794}}
  (\bibinfo{year}{2022}).
\newblock


\bibitem[Izacard et~al\mbox{.}(2022)]%
        {izacard2022unsupervised}
\bibfield{author}{\bibinfo{person}{Gautier Izacard}, \bibinfo{person}{Mathilde
  Caron}, \bibinfo{person}{Lucas Hosseini}, \bibinfo{person}{Sebastian Riedel},
  \bibinfo{person}{Piotr Bojanowski}, \bibinfo{person}{Armand Joulin}, {and}
  \bibinfo{person}{Edouard Grave}.} \bibinfo{year}{2022}\natexlab{}.
\newblock \showarticletitle{Unsupervised Dense Information Retrieval with
  Contrastive Learning}.
\newblock \bibinfo{journal}{\emph{TMLR}} (\bibinfo{year}{2022}).
\newblock
\showISSN{2835-8856}


\bibitem[Kan et~al\mbox{.}(2022)]%
        {kan2022fbnetgen}
\bibfield{author}{\bibinfo{person}{Xuan Kan}, \bibinfo{person}{Hejie Cui},
  \bibinfo{person}{Joshua Lukemire}, \bibinfo{person}{Ying Guo}, {and}
  \bibinfo{person}{Carl Yang}.} \bibinfo{year}{2022}\natexlab{}.
\newblock \showarticletitle{Fbnetgen: Task-aware gnn-based fmri analysis via
  functional brain network generation}. In \bibinfo{booktitle}{\emph{MIDL}}.
\newblock


\bibitem[Karpukhin et~al\mbox{.}(2020)]%
        {dpr}
\bibfield{author}{\bibinfo{person}{Vladimir Karpukhin}, \bibinfo{person}{Barlas
  Oguz}, \bibinfo{person}{Sewon Min}, \bibinfo{person}{Patrick Lewis},
  \bibinfo{person}{Ledell Wu}, \bibinfo{person}{Sergey Edunov},
  \bibinfo{person}{Danqi Chen}, {and} \bibinfo{person}{Wen-tau Yih}.}
  \bibinfo{year}{2020}\natexlab{}.
\newblock \showarticletitle{Dense Passage Retrieval for Open-Domain Question
  Answering}. In \bibinfo{booktitle}{\emph{EMNLP}}.
  \bibinfo{pages}{6769--6781}.
\newblock


\bibitem[Meng et~al\mbox{.}(2018)]%
        {meng2018weakly}
\bibfield{author}{\bibinfo{person}{Yu Meng}, \bibinfo{person}{Jiaming Shen},
  \bibinfo{person}{Chao Zhang}, {and} \bibinfo{person}{Jiawei Han}.}
  \bibinfo{year}{2018}\natexlab{}.
\newblock \showarticletitle{Weakly-supervised neural text classification}. In
  \bibinfo{booktitle}{\emph{CIKM}}. \bibinfo{pages}{983--992}.
\newblock


\bibitem[Meng et~al\mbox{.}(2020)]%
        {lotclass}
\bibfield{author}{\bibinfo{person}{Yu Meng}, \bibinfo{person}{Yunyi Zhang},
  \bibinfo{person}{Jiaxin Huang}, \bibinfo{person}{Chenyan Xiong},
  \bibinfo{person}{Heng Ji}, \bibinfo{person}{Chao Zhang}, {and}
  \bibinfo{person}{Jiawei Han}.} \bibinfo{year}{2020}\natexlab{}.
\newblock \showarticletitle{Text classification using label names only: A
  language model self-training approach}.
\newblock \bibinfo{journal}{\emph{EMNLP}} (\bibinfo{year}{2020}).
\newblock


\bibitem[Reimers and Gurevych(2019)]%
        {sentencebert}
\bibfield{author}{\bibinfo{person}{Nils Reimers} {and} \bibinfo{person}{Iryna
  Gurevych}.} \bibinfo{year}{2019}\natexlab{}.
\newblock \showarticletitle{Sentence-BERT: Sentence Embeddings using Siamese
  BERT-Networks}. In \bibinfo{booktitle}{\emph{EMNLP-IJCNLP}}.
  \bibinfo{pages}{3982--3992}.
\newblock


\bibitem[Robertson and Zaragoza(2009)]%
        {bm25}
\bibfield{author}{\bibinfo{person}{Stephen Robertson} {and}
  \bibinfo{person}{Hugo Zaragoza}.} \bibinfo{year}{2009}\natexlab{}.
\newblock \showarticletitle{The probabilistic relevance framework: BM25 and
  beyond}.
\newblock \bibinfo{journal}{\emph{Foundations and Trends in Information
  Retrieval}} \bibinfo{volume}{3}, \bibinfo{number}{4} (\bibinfo{year}{2009}),
  \bibinfo{pages}{333--389}.
\newblock


\bibitem[Trstenjak et~al\mbox{.}(2014)]%
        {tfidf}
\bibfield{author}{\bibinfo{person}{Bruno Trstenjak}, \bibinfo{person}{Sasa
  Mikac}, {and} \bibinfo{person}{Dzenana Donko}.}
  \bibinfo{year}{2014}\natexlab{}.
\newblock \showarticletitle{KNN with TF-IDF based framework for text
  categorization}.
\newblock \bibinfo{journal}{\emph{Procedia Engineering}}  \bibinfo{volume}{69}
  (\bibinfo{year}{2014}), \bibinfo{pages}{1356--1364}.
\newblock


\bibitem[Wang and Isola(2020)]%
        {wang2020understanding}
\bibfield{author}{\bibinfo{person}{Tongzhou Wang} {and}
  \bibinfo{person}{Phillip Isola}.} \bibinfo{year}{2020}\natexlab{}.
\newblock \showarticletitle{Understanding contrastive representation learning
  through alignment and uniformity on the hypersphere}. In
  \bibinfo{booktitle}{\emph{ICML}}.
\newblock


\bibitem[Wang et~al\mbox{.}(2021)]%
        {xclass}
\bibfield{author}{\bibinfo{person}{Zihan Wang}, \bibinfo{person}{Dheeraj
  Mekala}, {and} \bibinfo{person}{Jingbo Shang}.}
  \bibinfo{year}{2021}\natexlab{}.
\newblock \showarticletitle{{X}-Class: Text Classification with Extremely Weak
  Supervision}. In \bibinfo{booktitle}{\emph{NAACL}}.
  \bibinfo{pages}{3043--3053}.
\newblock


\bibitem[Wang et~al\mbox{.}(2023)]%
        {wang2023benchmark}
\bibfield{author}{\bibinfo{person}{Zihan Wang}, \bibinfo{person}{Tianle Wang},
  \bibinfo{person}{Dheeraj Mekala}, {and} \bibinfo{person}{Jingbo Shang}.}
  \bibinfo{year}{2023}\natexlab{}.
\newblock \showarticletitle{A Benchmark on Extremely Weakly Supervised Text
  Classification: Reconcile Seed Matching and Prompting Approaches}.
\newblock \bibinfo{journal}{\emph{arXiv preprint arXiv:2305.12749}}
  (\bibinfo{year}{2023}).
\newblock


\bibitem[Xia et~al\mbox{.}(2022)]%
        {xia2022fastclass}
\bibfield{author}{\bibinfo{person}{Tingyu Xia}, \bibinfo{person}{Yue Wang},
  \bibinfo{person}{Yuan Tian}, {and} \bibinfo{person}{Yi Chang}.}
  \bibinfo{year}{2022}\natexlab{}.
\newblock \showarticletitle{FastClass: A Time-Efficient Approach to
  Weakly-Supervised Text Classification}.
\newblock \bibinfo{journal}{\emph{EMNLP}} (\bibinfo{year}{2022}).
\newblock


\bibitem[Xie et~al\mbox{.}(2021)]%
        {xie2021learning}
\bibfield{author}{\bibinfo{person}{Yi Xie}, \bibinfo{person}{Yuqing Sun}, {and}
  \bibinfo{person}{Elisa Bertino}.} \bibinfo{year}{2021}\natexlab{}.
\newblock \showarticletitle{Learning domain semantics and cross-domain
  correlations for paper recommendation}. In \bibinfo{booktitle}{\emph{SIGIR}}.
  \bibinfo{pages}{706--715}.
\newblock


\bibitem[Xu et~al\mbox{.}(2023)]%
        {xu2023neighborhood}
\bibfield{author}{\bibinfo{person}{Ran Xu}, \bibinfo{person}{Yue Yu},
  \bibinfo{person}{Hejie Cui}, \bibinfo{person}{Xuan Kan},
  \bibinfo{person}{Yanqiao Zhu}, \bibinfo{person}{Joyce Ho},
  \bibinfo{person}{Chao Zhang}, {and} \bibinfo{person}{Carl Yang}.}
  \bibinfo{year}{2023}\natexlab{}.
\newblock \showarticletitle{Neighborhood-Regularized Self-Training for Learning
  with Few Labels}. In \bibinfo{booktitle}{\emph{AAAI}},
  Vol.~\bibinfo{volume}{37}.
\newblock


\bibitem[Xu et~al\mbox{.}(2022)]%
        {xu2022counterfactual}
\bibfield{author}{\bibinfo{person}{R. Xu}, \bibinfo{person}{Y. Yu},
  \bibinfo{person}{C. Zhang}, \bibinfo{person}{M.~K Ali}, \bibinfo{person}{JC.
  Ho}, {and} \bibinfo{person}{C. Yang}.} \bibinfo{year}{2022}\natexlab{}.
\newblock \showarticletitle{Counterfactual and factual reasoning over
  hypergraphs for interpretable clinical predictions on ehr}. In
  \bibinfo{booktitle}{\emph{Machine Learning for Health}}. PMLR,
  \bibinfo{pages}{259--278}.
\newblock


\bibitem[Yang et~al\mbox{.}(2022)]%
        {yang2022pre}
\bibfield{author}{\bibinfo{person}{Yi Yang}, \bibinfo{person}{Hejie Cui}, {and}
  \bibinfo{person}{Carl Yang}.} \bibinfo{year}{2022}\natexlab{}.
\newblock \showarticletitle{Pre-train Graph Neural Networks for Brain Network
  Analysis}. In \bibinfo{booktitle}{\emph{IEEE-Big Data}}.
\newblock


\bibitem[Yu et~al\mbox{.}(2022a)]%
        {yu-etal-2022-actune}
\bibfield{author}{\bibinfo{person}{Yue Yu}, \bibinfo{person}{Lingkai Kong},
  \bibinfo{person}{Jieyu Zhang}, \bibinfo{person}{Rongzhi Zhang}, {and}
  \bibinfo{person}{Chao Zhang}.} \bibinfo{year}{2022}\natexlab{a}.
\newblock \showarticletitle{{A}c{T}une: Uncertainty-Based Active Self-Training
  for Active Fine-Tuning of Pretrained Language Models}. In
  \bibinfo{booktitle}{\emph{NAACL}}. \bibinfo{pages}{1422--1436}.
\newblock


\bibitem[Yu et~al\mbox{.}(2022b)]%
        {yu2022coco}
\bibfield{author}{\bibinfo{person}{Yue Yu}, \bibinfo{person}{Chenyan Xiong},
  \bibinfo{person}{Si Sun}, \bibinfo{person}{Chao Zhang}, {and}
  \bibinfo{person}{Arnold Overwijk}.} \bibinfo{year}{2022}\natexlab{b}.
\newblock \showarticletitle{COCO-DR: Combating Distribution Shifts in Zero-Shot
  Dense Retrieval with Contrastive and Distributionally Robust Learning}. In
  \bibinfo{booktitle}{\emph{EMNLP}}. \bibinfo{pages}{1462--1479}.
\newblock


\bibitem[Yu et~al\mbox{.}(2021)]%
        {yu-etal-2021-fine}
\bibfield{author}{\bibinfo{person}{Yue Yu}, \bibinfo{person}{Simiao Zuo},
  \bibinfo{person}{Haoming Jiang}, \bibinfo{person}{Wendi Ren},
  \bibinfo{person}{Tuo Zhao}, {and} \bibinfo{person}{Chao Zhang}.}
  \bibinfo{year}{2021}\natexlab{}.
\newblock \showarticletitle{Fine-Tuning Pre-trained Language Model with Weak
  Supervision: A Contrastive-Regularized Self-Training Approach}. In
  \bibinfo{booktitle}{\emph{NAACL}}. \bibinfo{pages}{1063--1077}.
\newblock


\bibitem[Zhang et~al\mbox{.}(2021)]%
        {zhang2wrench}
\bibfield{author}{\bibinfo{person}{Jieyu Zhang}, \bibinfo{person}{Yue Yu},
  \bibinfo{person}{Yinghao Li}, \bibinfo{person}{Yujing Wang},
  \bibinfo{person}{Yaming Yang}, \bibinfo{person}{Mao Yang}, {and}
  \bibinfo{person}{Alexander Ratner}.} \bibinfo{year}{2021}\natexlab{}.
\newblock \showarticletitle{WRENCH: A Comprehensive Benchmark for Weak
  Supervision}. In \bibinfo{booktitle}{\emph{NeurIPS}}.
\newblock


\bibitem[Zhang et~al\mbox{.}(2022b)]%
        {zhang2022adaptive}
\bibfield{author}{\bibinfo{person}{Rongzhi Zhang}, \bibinfo{person}{Rebecca
  West}, \bibinfo{person}{Xiquan Cui}, {and} \bibinfo{person}{Chao Zhang}.}
  \bibinfo{year}{2022}\natexlab{b}.
\newblock \showarticletitle{Adaptive Multi-view Rule Discovery for
  Weakly-Supervised Compatible Products Prediction}. In
  \bibinfo{booktitle}{\emph{KDD}}. \bibinfo{pages}{4521--4529}.
\newblock


\bibitem[Zhang et~al\mbox{.}(2022c)]%
        {prboost}
\bibfield{author}{\bibinfo{person}{Rongzhi Zhang}, \bibinfo{person}{Yue Yu},
  \bibinfo{person}{Pranav Shetty}, \bibinfo{person}{Le Song}, {and}
  \bibinfo{person}{Chao Zhang}.} \bibinfo{year}{2022}\natexlab{c}.
\newblock \showarticletitle{PRBoost: Prompt-Based Rule Discovery and Boosting
  for Interactive Weakly-Supervised Learning}. In
  \bibinfo{booktitle}{\emph{ACL}}.
\newblock


\bibitem[Zhang et~al\mbox{.}(2015)]%
        {Zhang2015CharacterlevelCN}
\bibfield{author}{\bibinfo{person}{Xiang Zhang}, \bibinfo{person}{Junbo~Jake
  Zhao}, {and} \bibinfo{person}{Yann LeCun}.} \bibinfo{year}{2015}\natexlab{}.
\newblock \showarticletitle{Character-level Convolutional Networks for Text
  Classification}. In \bibinfo{booktitle}{\emph{NIPS}}.
\newblock


\bibitem[Zhang et~al\mbox{.}(2023a)]%
        {zhang2023pre}
\bibfield{author}{\bibinfo{person}{Yu Zhang}, \bibinfo{person}{Hao Cheng},
  \bibinfo{person}{Zhihong Shen}, \bibinfo{person}{Xiaodong Liu},
  \bibinfo{person}{Ye-Yi Wang}, {and} \bibinfo{person}{Jianfeng Gao}.}
  \bibinfo{year}{2023}\natexlab{a}.
\newblock \showarticletitle{Pre-training Multi-task Contrastive Learning Models
  for Scientific Literature Understanding}.
\newblock \bibinfo{journal}{\emph{arXiv preprint arXiv:2305.14232}}
  (\bibinfo{year}{2023}).
\newblock


\bibitem[Zhang et~al\mbox{.}(2023b)]%
        {zhang2023effect}
\bibfield{author}{\bibinfo{person}{Yu Zhang}, \bibinfo{person}{Bowen Jin},
  \bibinfo{person}{Qi Zhu}, \bibinfo{person}{Yu Meng}, {and}
  \bibinfo{person}{Jiawei Han}.} \bibinfo{year}{2023}\natexlab{b}.
\newblock \showarticletitle{The Effect of Metadata on Scientific Literature
  Tagging: A Cross-Field Cross-Model Study}. In
  \bibinfo{booktitle}{\emph{WWW}}. \bibinfo{pages}{1626--1637}.
\newblock


\bibitem[Zhang et~al\mbox{.}(2022a)]%
        {zhang2022seed}
\bibfield{author}{\bibinfo{person}{Yu Zhang}, \bibinfo{person}{Yu Meng},
  \bibinfo{person}{Xuan Wang}, \bibinfo{person}{Sheng Wang}, {and}
  \bibinfo{person}{Jiawei Han}.} \bibinfo{year}{2022}\natexlab{a}.
\newblock \showarticletitle{Seed-Guided Topic Discovery with Out-of-Vocabulary
  Seeds}. In \bibinfo{booktitle}{\emph{NAACL}}. \bibinfo{pages}{279--290}.
\newblock


\bibitem[Zhu et~al\mbox{.}(2022)]%
        {zhu2022structure}
\bibfield{author}{\bibinfo{person}{Yanqiao Zhu}, \bibinfo{person}{Yichen Xu},
  \bibinfo{person}{Hejie Cui}, \bibinfo{person}{Carl Yang},
  \bibinfo{person}{Qiang Liu}, {and} \bibinfo{person}{Shu Wu}.}
  \bibinfo{year}{2022}\natexlab{}.
\newblock \showarticletitle{Structure-enhanced heterogeneous graph contrastive
  learning}. In \bibinfo{booktitle}{\emph{SDM}}.
\newblock


\bibitem[Zhuang et~al\mbox{.}(2022)]%
        {zhuang-etal-2022-resel}
\bibfield{author}{\bibinfo{person}{Yuchen Zhuang}, \bibinfo{person}{Yinghao
  Li}, \bibinfo{person}{Junyang Zhang}, \bibinfo{person}{Yue Yu},
  \bibinfo{person}{Yingjun Mou}, \bibinfo{person}{Xiang Chen},
  \bibinfo{person}{Le Song}, {and} \bibinfo{person}{Chao Zhang}.}
  \bibinfo{year}{2022}\natexlab{}.
\newblock \showarticletitle{{R}e{S}el: N-ary Relation Extraction from
  Scientific Text and Tables by Learning to Retrieve and Select}. In
  \bibinfo{booktitle}{\emph{EMNLP}}. \bibinfo{pages}{730--744}.
\newblock


\end{thebibliography}

\end{document}